# Quantum-Inspired Harmonic Decision Models: A Computational Framework for Music Generation

**Josef Pavlíček[1], Petra Pavlíčková[1], Martin Molhanec[2]**

[1] CTU, Faculty of Information Technology, Thákurova 9, Prague 6, 160 00, Czech Republic

[2] CTU, Faculty of Electrical Engineering, CTU, Technická 2, Prague 6, 160 00, Czech Republic

**ABSTRACT**

This paper introduces a quantum-inspired computational framework for harmonic decision-making in music. The proposed approach formulates harmonization as an optimization problem within a structured combinatorial space, where multiple candidate chord sequences are evaluated under interacting musical constraints. The model combines an interference-based harmonization stage with a classical optimization procedure grounded in tonal harmony. The quantum-inspired component enables the parallel consideration of multiple harmonic alternatives, while the classical stage refines the resulting sequences to ensure structural coherence and stylistic plausibility.

The framework is evaluated on selected musical examples, including Autumn Leaves and It's a Long Way to Tipperary. Quantitative analysis shows that the optimization stage significantly reduces chord density, increases harmonic stability, and improves functional organization. At the same time, expert evaluation highlights the importance of stylistic context, demonstrating that increased harmonic complexity is not always perceived as more natural. The results suggest that harmonic generation can be interpreted as a structured decision-making process in a constrained search space. The proposed approach provides a computational model that integrates domain-specific knowledge with an interference-based search mechanism.

Although preliminary, this work indicates that quantum-inspired methods may offer a useful framework for modeling complex decision processes in creative domains such as music. The proposed framework contributes to ongoing research on quantum-inspired models of cognition and decision-making in complex biological and creative systems.

# 1 Introduction

Understanding how humans generate structured creative decisions remains an open question (Simon, 1972; Gigerenzer and Gaissmaier, 2011). This question spans multiple scientific disciplines, including cognitive science, artificial intelligence, neuroscience, and music theory (Bharucha and Krumhansl, 1983). Musical composition provides a particularly suitable domain for investigating this problem, as it combines strict formal rules with a high degree of creative freedom (Krumhansl and Kessler, 1982). Composers operate within well-defined—though often implicitly learned—theoretical frameworks such as tonal harmony, voice leading, and functional relationships between chords (Lerdahl and Jackendoff, 1983), yet the resulting musical structures often appear as spontaneous creative insights rather than deterministic outputs of algorithmic procedures.

Recent advances in artificial intelligence have led to significant progress in automatic music generation (Briot et al., 2019; Huang et al., 2018). Modern systems based on deep learning and large-scale statistical models are capable of producing stylistically convincing musical outputs by learning patterns from extensive datasets. However, these approaches primarily rely on statistical imitation and probabilistic prediction. In this sense, it remains an open question to what extent such predictions can be considered "intelligent" in a human decision-making sense. As a consequence, these models tend to reproduce stylistic regularities present in training data rather than explicitly modeling the structured decision processes underlying harmonic reasoning.

In contrast, human composers navigate a complex space of possible harmonic structures while respecting theoretical constraints such as tonal function, melodic compatibility, and stylistic coherence. This observation suggests that musical composition can be interpreted as a decision process within a highly structured combinatorial space. Each melodic fragment can be harmonized by many possible chord sequences; however, only a small subset satisfies musical constraints while simultaneously producing aesthetically convincing results.

Importantly, the notion of aesthetic value is not static but evolves over time. For example, the tritone interval was historically considered undesirable in medieval music ("diabolus in musica"), whereas in contemporary music it is widely accepted and even characteristic of certain genres, such as heavy metal. This suggests that the evaluation of harmonic correctness is context-dependent and influenced by style, historical development, and listener perception.

Exploring such spaces efficiently therefore represents a non-trivial computational problem. Classical optimization approaches, including rule-based systems and dynamic programming, can capture certain aspects of harmonic reasoning; however, they often require extensive heuristics to avoid combinatorial explosion.

In recent years, quantum-inspired models have been proposed in various areas of cognition and decision-making (Khrennikov, 2010, 2017; Busemeyer and Bruza, 2012). These models do not assume that biological systems perform physical quantum computation. Instead, they employ mathematical structures derived from quantum theory—such as superposition, interference, and probabilistic amplitude dynamics—to describe complex decision processes. Such approaches have proven useful in modeling context-dependent decision-making, ambiguity resolution, and preference shifts.

Motivated by these approaches, this work investigates a quantum-inspired computational framework for harmonic decision-making in music. The central idea is to formulate the selection of harmonic progressions as an optimization problem within a structured space of chord configurations. Instead of relying on purely deterministic search, the proposed model employs an interference-based mechanism inspired by the Quantum Approximate Optimization Algorithm (QAOA), enabling parallel evaluation of multiple harmonic candidates.

The proposed framework consists of two complementary components. First, a quantum-inspired harmonization stage generates candidate chord structures through interference-based exploration of a discrete harmonic state space. Second, a classical optimization stage refines these candidates using established principles of tonal harmony, including functional relationships, chord inversions, and voice-leading constraints. This hybrid architecture integrates structured musical knowledge with an efficient search mechanism.

Beyond music generation, the proposed model can be interpreted as an experimental platform for studying structured decision-making processes in creative cognition. By analyzing harmonizations generated under different optimization strategies, it becomes possible to investigate whether interference-based search mechanisms provide advantages over classical approaches when navigating complex creative spaces.

The remainder of this paper is organized as follows. Section 2 introduces the theoretical background. Section 3 formalizes the harmonic search space. Section 4 presents the quantum-inspired harmonization algorithm. Section 5 describes the classical optimization stage. Section 6 provides experimental evaluation. Finally, Section 7 discusses implications for computational creativity and decision-making.

# 2 Background

This section introduces the theoretical foundations underlying the proposed approach. We first outline tonal harmony as a structured system of constraints, then discuss decision-making in complex combinatorial spaces, and finally review relevant concepts from quantum-inspired models of cognition.

## 2.1 Tonal Harmony as a Constraint System

Tonal harmony can be understood as a structured system of constraints governing the relationships between musical elements. Within this framework, chords are not independent entities but are organized according to functional roles, such as tonic, dominant, and subdominant. These roles define expectations and permissible transitions, forming a network of dependencies that guides harmonic progression.

In addition to functional relationships, harmonic structures are constrained by voice-leading principles, which impose local continuity conditions on individual melodic lines. These constraints limit abrupt changes and favor smooth transitions between successive chords. As a result, the space of possible harmonizations for a given melody is highly structured but non-trivial.

Importantly, this system does not uniquely determine a single solution. Multiple harmonic realizations may satisfy the same set of constraints while differing in stylistic or expressive qualities. This inherent multiplicity suggests that harmonic composition involves selecting among competing alternatives rather than deriving a single deterministic outcome.

## 2.2 Decision-Making in Structured Spaces

The process of selecting harmonic progressions can be interpreted as decision-making within a structured combinatorial space. Each candidate solution corresponds to a sequence of chords that must satisfy multiple interacting constraints, including tonal function, melodic compatibility, and stylistic coherence.

Such spaces are typically characterized by a large number of possible configurations, most of which violate one or more constraints. Identifying valid and musically meaningful solutions therefore requires efficient exploration strategies. Classical approaches to this problem often rely on rule-based systems, heuristic search, or optimization techniques such as dynamic programming.

However, these methods face limitations when dealing with highly complex or context-dependent decision spaces. In particular, they may struggle to balance competing constraints or to capture the flexibility observed in human creative behavior. This has motivated the exploration of alternative computational paradigms capable of representing and navigating such spaces more effectively.

### 2.3 Quantum-Inspired Models of Cognition

Quantum-inspired models of cognition provide a mathematical framework for describing decision-making processes that deviate from classical probabilistic reasoning (Khrennikov, 2010, 2013; Busemeyer and Bruza, 2012). Rather than assuming that decisions arise from fixed probability distributions, these models represent cognitive states as superpositions of multiple potential outcomes.

Within this framework, the act of decision-making can be interpreted as a transformation of these states, where interference effects influence the likelihood of particular outcomes. This allows the model to capture context-dependent behavior, ambiguity resolution, and non-classical probability patterns observed in empirical studies. Similar approaches have also been explored in the context of biological and cognitive systems, where quantum-like models have been used to describe decision processes under uncertainty and contextual influence (Asano et al., 2011).

Importantly, the use of quantum formalism in this context does not imply that cognitive processes are physically quantum. Instead, it provides a flexible mathematical language for modeling complex systems in which multiple competing possibilities coexist and interact.

These properties make quantum-inspired approaches particularly relevant for domains such as musical harmony, where multiple valid solutions exist and must be evaluated under interacting constraints.

## 3 Harmonic Search Space

In this section, we formalize harmonic generation as a search process within a structured space of chord configurations. The goal is to define the elements of this space, their formal representation, and the constraints that govern valid harmonic solutions.

The proposed formulation builds on established formal approaches to harmony, where chords are represented as structured sets of pitch classes and constructed through interval-based transformations.

### 3.1 Melody Representation

Let $M = (m_1, m_2, ..., m_T)$ denote a melodic sequence of length T, where each element $m_t$ represents a pitch event at time step t.

We define the pitch space as the set of pitch classes modulo 12:

$$\mathcal{P} = \mathbb{Z}_{12} = \{0, 1, ..., 11\}$$

where each melodic element $m_t \in \mathcal{P}$ (or a higher-level representation such as MIDI pitch mapped to pitch class).

For the purposes of harmonic generation, the melody defines a sequence of anchor points that constrain the selection of compatible chords. At each time step t, the corresponding harmonic choice must be consistent with the melodic note $m_t$, either by containing it as a chord tone or by supporting it as a non-chord tone under standard harmonic conventions.

### 3.2 Chord Candidate Space

Let $C_t$ denote the set of candidate chords available at time step t.

Each chord $c \in C_t$ is represented as a finite subset of pitch classes:

$$c \subseteq \mathcal{P}.$$

In addition to its pitch-class content, each chord carries structural attributes such as chord quality (e.g., major, minor, dominant seventh) and inversion.

Chord construction can be interpreted as the application of interval-based transformations to a root pitch. Formally, given a root $r \in \mathcal{P}$, a chord can be constructed as:

$$c = \{r, r + \Delta_1, r + \Delta_2, ...\} \bmod 12,$$

where $\Delta_i$ represent interval offsets (e.g., major third, perfect fifth, minor seventh).

The global harmonic configuration is defined as a sequence:

$$C = (c_1, c_2, ..., c_T),$$

where $c_t \in C_t$ for each time step t.

The size of the search space grows exponentially with T, as the total number of possible harmonic sequences is given by the Cartesian product:

$$|C_1 \times C_2 \times \ldots \times C_T|.$$

This combinatorial growth makes exhaustive search impractical even for relatively short melodies, motivating the need for structured search strategies.

### 3.3 Constraint System

The set of valid harmonic sequences is restricted by a system of constraints reflecting principles of tonal harmony. These constraints define a structured subspace of musically plausible solutions.

They can be divided into three main categories:

(i) **Functional constraints:**
Chord progressions should follow plausible functional relationships, such as transitions between tonic, dominant, and subdominant regions.

(ii) **Melodic compatibility constraints:**
Each chord $c_t$ must be compatible with the corresponding melodic element $m_t$, ensuring that the melody is harmonically supported.

(iii) **Voice-leading constraints:**
Transitions between consecutive chords $(c_t, c_{t+1})$ should minimize abrupt changes and favor smooth movement between individual voices.

These constraints significantly reduce the effective search space; however, even within this restricted subset, multiple competing harmonic configurations typically exist.

Therefore, harmonic generation can be formulated as an optimization problem over a constrained combinatorial space, where the objective is to identify sequences that best satisfy a weighted combination of these criteria.

## 4 Quantum-Inspired Harmonizer

In this section, we introduce a quantum-inspired approach to harmonic generation, formulated as a structured search process over the combinatorial space defined in Section 3.

The proposed method does not assume that physical quantum processes are involved in musical cognition. Instead, it adopts mathematical principles inspired by quantum theory—particularly superposition and interference—to guide the exploration of multiple harmonic candidates in parallel.

### 4.1 Superposition of Harmonic Candidates

At each time step t, instead of selecting a single chord $c_t \in C_t$ deterministically, we consider a weighted representation over multiple candidates.

This can be interpreted as a superposition-like state over the candidate set:

$$\psi_t = \sum_{c \in C_t} w(c) \cdot |c\rangle,$$

where w(c) represents a weight associated with chord candidate c, and $|c\rangle$ denotes an abstract representation of that chord.

Unlike physical quantum states, these weights do not need to satisfy strict normalization constraints; rather, they serve as a flexible mechanism for representing competing harmonic hypotheses.

### 4.2 Interference-Based Evaluation

The key idea of the proposed model is that harmonic candidates are not evaluated independently. Instead, their contributions interact through an interference-like mechanism that reflects how different constraints reinforce or suppress specific harmonic configurations.

Each candidate chord c is assigned a score based on multiple criteria:

- melodic compatibility,
- functional coherence,
- voice-leading smoothness.

These criteria can be combined into a global evaluation function:

$$E(c_1, c_2, ..., c_T),$$

which assigns a scalar value to a complete harmonic sequence.

Rather than evaluating each sequence independently, the model iteratively adjusts candidate weights such that harmonically consistent configurations are reinforced, while inconsistent ones are suppressed. This process is conceptually analogous to constructive and destructive interference.

### 4.3 Iterative Refinement Process

The harmonization process proceeds iteratively.

Starting from an initial distribution of candidate chords at each time step, the model repeatedly updates the weights w(c) based on their contribution to globally coherent harmonic sequences.

At each iteration, the following steps are performed (Figure 1):

- Generate candidate combinations of chords across time steps.
- Evaluate these combinations using a global constraint-based scoring function E, which quantifies their melodic compatibility, functional coherence, and voice-leading smoothness.
- Update the weights w(c) to favour candidates that participate in higher-scoring configurations.

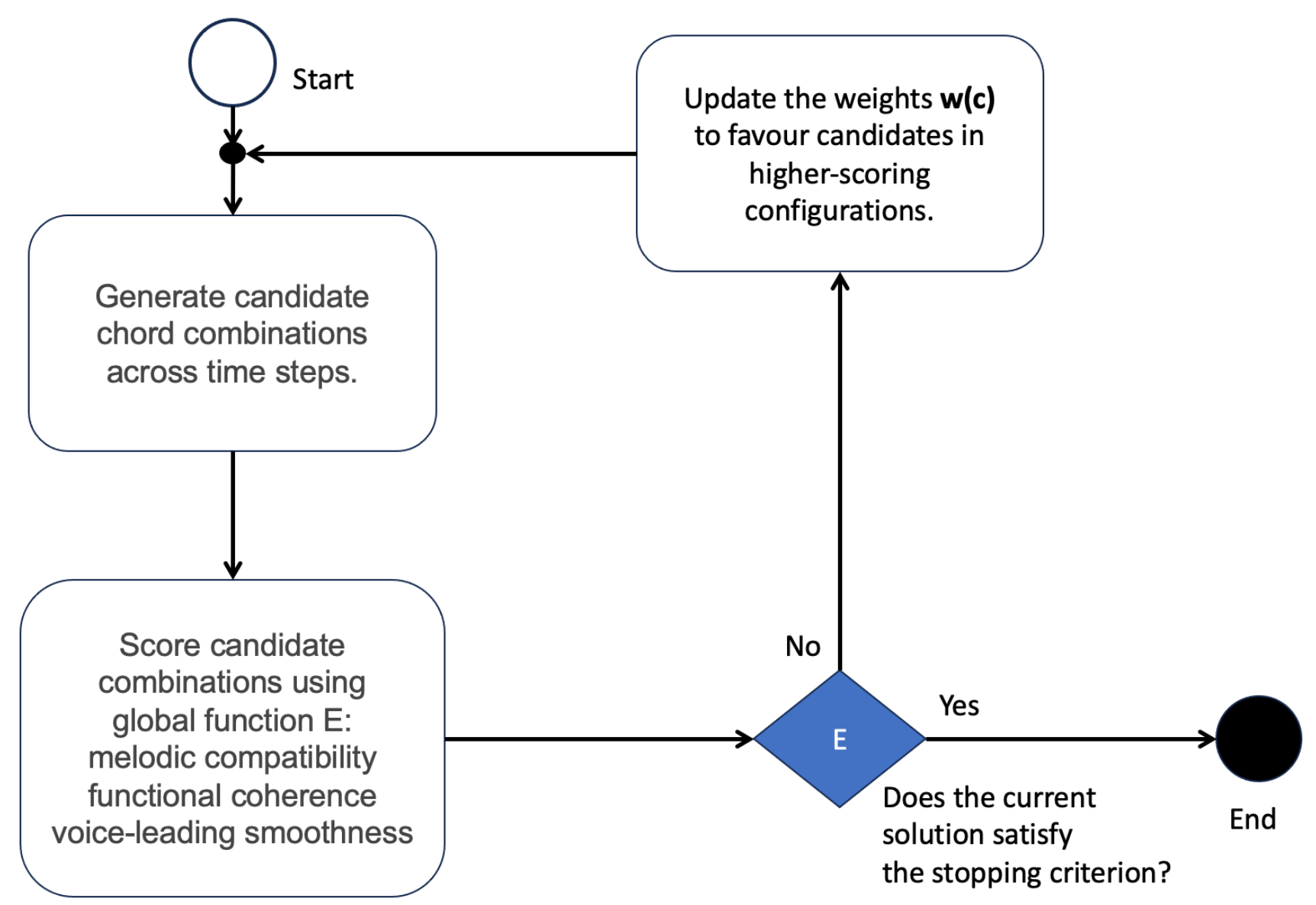

**Figure 1**. Quantum harmonic optimization flowchart (source: Author)

This iterative refinement gradually concentrates the search toward a subset of harmonically consistent solutions.

### 4.4 Relation to Approximate Optimization

The proposed approach can be interpreted as a heuristic approximation to combinatorial optimization methods. In particular, it shares conceptual similarities with algorithms that iteratively refine candidate solutions using feedback from a global objective function.

While inspired by the Quantum Approximate Optimization Algorithm (QAOA), the present model remains entirely classical in its implementation. The reference to quantum-inspired mechanisms serves primarily as a conceptual framework for understanding how multiple candidate solutions can be explored and selectively amplified.

## 5 Hybrid Classical Optimization

While the quantum-inspired harmonization stage provides a mechanism for exploring the space of possible harmonic configurations, the resulting candidates may still contain local inconsistencies or suboptimal transitions.

To address this, we introduce a subsequent classical optimization stage that refines the harmonic sequence using established principles of tonal harmony.

This hybrid approach combines exploratory search with rule-based correction, allowing the system to balance global coherence with local musical correctness.

## 5.1 Harmonic Smoothing

The first refinement step focuses on harmonic smoothing, aiming to eliminate abrupt or stylistically implausible transitions between consecutive chords.

Given a sequence $C = (c_1, c_2, ..., c_T)$, we evaluate each pair $(c_t, c_{t+1})$ and apply local adjustments where necessary. These adjustments may include:

- replacing a chord with a functionally related alternative,
- inserting intermediate chords to improve continuity,
- resolving unstable harmonic movements.

This process improves the overall flow of the harmonic progression while preserving compatibility with the melody.

## 5.2 Inversion Selection

Next, chord inversions are selected to optimize voice leading and reduce large pitch movements between successive chords.

For each chord $c_t$, multiple inversion variants are considered. The selection criterion prioritizes:

- minimal total voice displacement,
- avoidance of parallel motion where stylistically inappropriate,
- preservation of chord identity and function.

This step transforms abstract chord representations into more realistic harmonic realizations.

## 5.3 Cadence Stabilization

The final refinement stage addresses cadence stabilization, ensuring that the harmonic sequence exhibits structurally meaningful points of resolution.

Particular attention is given to phrase endings, where standard cadential patterns (e.g., dominant–tonic resolution) are preferred.

If necessary, the model adjusts the final chords in a segment to reinforce a sense of closure, while maintaining consistency with preceding harmonic context.

### 5.4 Integration with the Search Process

The classical optimization stage operates as a post-processing layer applied to the output of the quantum-inspired harmonizer.

Importantly, this stage does not replace the exploratory mechanism, but complements it. The harmonizer is responsible for navigating the global search space, while the classical optimizer enforces local consistency and stylistic correctness.

Together, these components form a hybrid architecture that integrates structured musical knowledge with an efficient search strategy over complex combinatorial spaces.

## 6 Experiments and Illustrative Evaluation

The aim of the experimental section is not to provide a large-scale benchmark, but rather to illustrate how the proposed framework behaves on concrete musical examples and whether its outputs exhibit musically plausible decision patterns.

Two types of harmonic outputs were considered:

- the direct output of the quantum-inspired harmonizer,
- the refined output after classical optimization.
- All outputs were generated from the same input melody and exported as MusicXML for inspection and comparison.

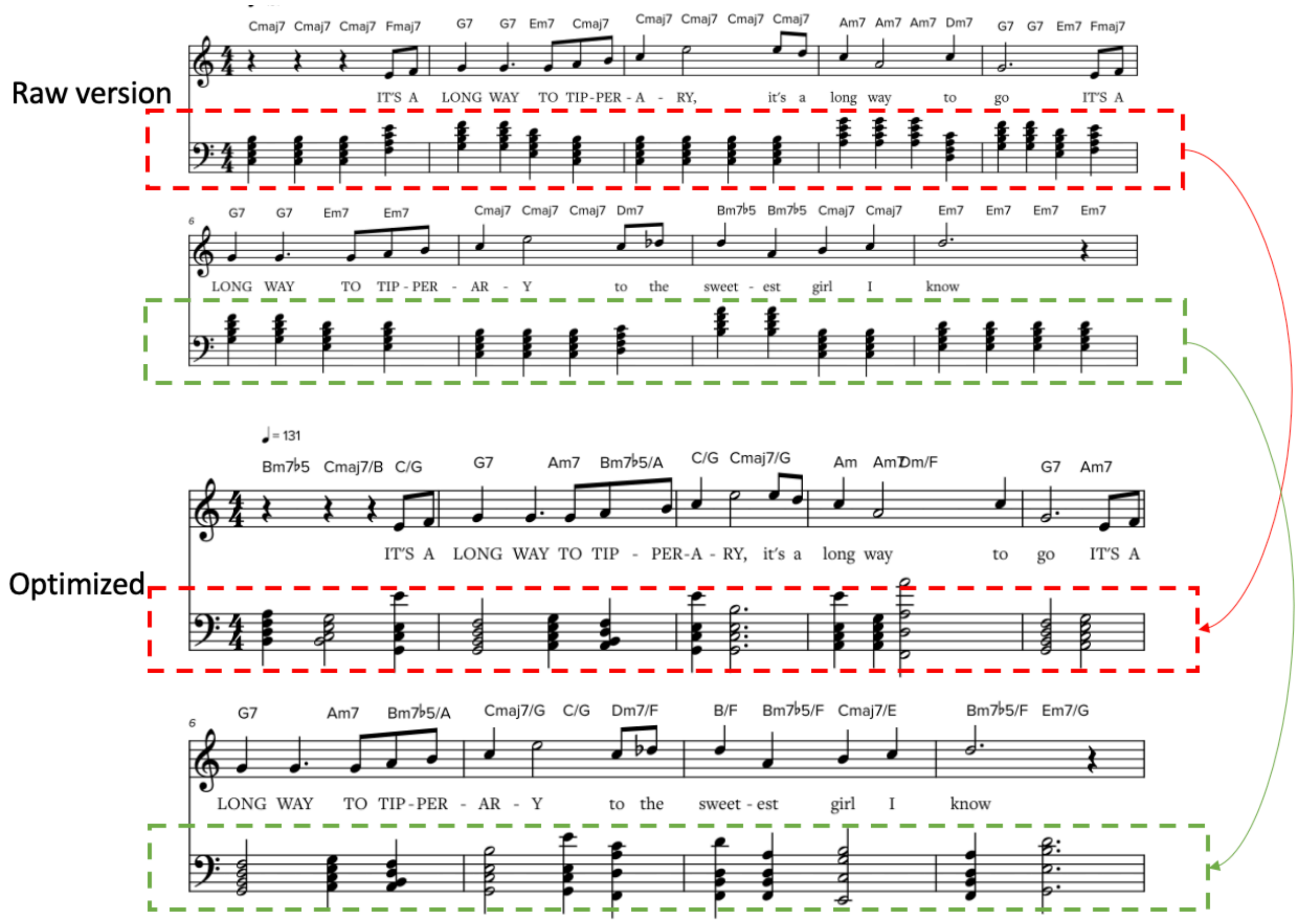

**Figure 2**. Tipperary 9 beats– raw vs optimized harmonization (source: Author)

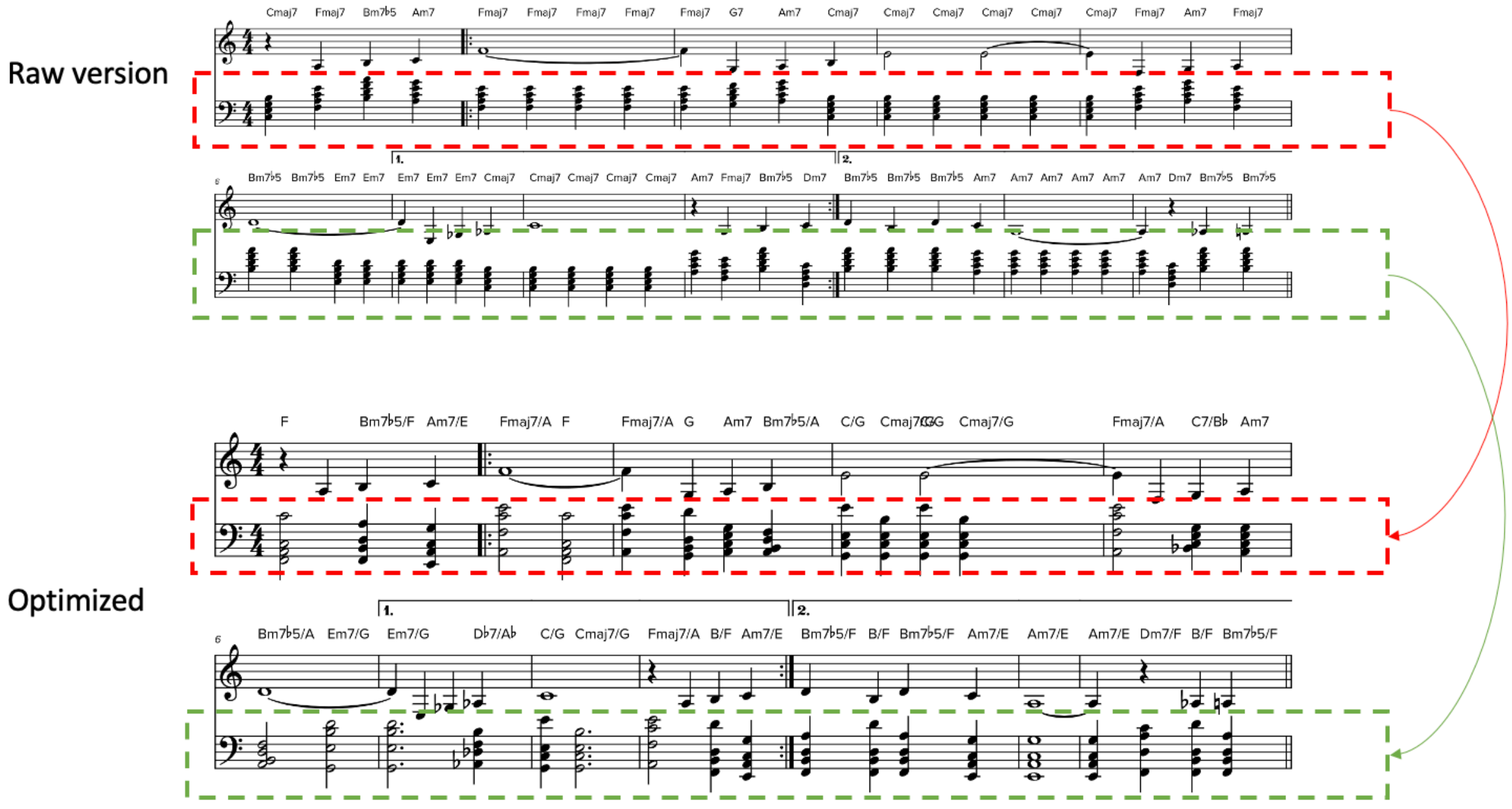

**Figure 3**. Autumn Leaves 8 beats – raw vs optimized harmonization (source: Author)

The generated harmonisations for both musical examples are illustrated in Figures 2 and 3. Each figure shows a comparison between the raw harmonizer output and the optimized version.

These examples provide a qualitative reference for interpreting the quantitative metrics presented in the following sections.

## 6.1 Evaluation Metrics

The generated harmonizations were evaluated using a combination of structural and expert-based criteria. Structurally, we considered melodic compatibility, functional coherence, and the density of harmonic changes. Melodic compatibility was defined as the proportion of time steps at which the melody note was contained in the corresponding chord. Functional coherence was assessed in terms of plausibility of transitions between tonic, subdominant, and dominant regions. In addition, the number of chord changes was used as a simple measure of harmonic stability versus fragmentation.

To complement these structural measures, the resulting harmonizations were also assessed qualitatively by a professional musician, who evaluated the outputs with respect to harmonic naturalness, stylistic plausibility, and the perceived quality of cadential resolution.

In addition to qualitative expert evaluation, several quantitative indicators were computed to characterize the generated harmonizations.

These include:

- Chord density (average number of chord changes per measure)
- Chord complexity (ratio of extended chords, e.g., seventh chords, to triads)
- Melodic coverage (percentage of melody notes contained in the corresponding chord)
- Harmonic stability (average duration of chord persistence)

## 6.2 Quantitative Analysis

To complement the qualitative evaluation, several quantitative indicators were computed for both the raw harmonizer output and the optimized version.

Table 1 summarizes the chord density, defined as the average number of chord changes per measure. The optimization stage leads to a substantial reduction in chord density for both evaluated melodies. In the case of Autumn Leaves, chord density decreases from 4.00 to 2.71 changes per measure, while for It’s a Long Way to Tipperary it decreases from 4.00 to 2.53. This corresponds to a reduction of 1.29 and 1.47 chord changes per measure, respectively, indicating a significant decrease in harmonic fragmentation.

**Table 1.** Chord density and reduction after optimization.

| Song | Raw Density | Optimized Density | Reduction |
|---|---|---|---|
| Autumn Leaves | 4,00 | 2,71 | 1,29 |
| It's a Long Way to Tipperary | 4,00 | 2,53 | 1,47 |

At the same time, chord duration increases notably. As shown in Table 3, the average chord duration nearly doubles in the optimized version (from 1.00 to 1.87 for Autumn Leaves and from 1.00 to 1.98 for Tipperary), suggesting improved harmonic stability and smoother structural flow.

Chord density (Table 1) further reflects this effect. While the raw harmonizer assigns chords at a high temporal resolution (often one chord per beat, primarily for technical and analytical purposes), the optimizer reduces the number of chord changes (from 4.00 to 2.71 for Autumn Leaves and from 4.00 to 2.53 for Tipperary), effectively simplifying the harmonic structure and extending chord durations.

The complexity of harmonic structures was assessed using the ratio of extended chords (e.g., seventh chords). The results (Table 2) indicate that the raw harmonizer produces a consistently high proportion of extended chords (up to 100%), whereas the optimized version reduces this ratio (to 77.6% and 74.4%). This reflects a shift toward a more balanced harmonic vocabulary, closer to stylistically plausible practice.

Melodic coverage, defined as the proportion of melody notes contained in the corresponding chord, shows a moderate decrease after optimization (e.g., from 96.2% to 84.6% for Autumn Leaves). This suggests that the optimizer allows for a more flexible harmonic interpretation, including non-chord tones and passing tones, rather than enforcing strict chord-tone alignment.

The distribution of harmonic functions (Table 2) reveals additional structural changes. While the raw harmonizer tends to favor tonic-related chords, the optimized version reduces their proportion (e.g., from 61.6% to 52.8% in Autumn Leaves) and increases the presence of dominant functions (from 21.4% to 23.2%). In Tipperary, dominant usage increases more substantially (from 14.7% to 23.5%). This indicates that the optimizer introduces greater harmonic tension and directional movement within the progression.

Finally, the average bass movement between consecutive chords was reduced ($\Delta = 0.56$ for Autumn Leaves and $\Delta = 0.72$ for Tipperary), indicating improved voice-leading smoothness and fewer abrupt transitions.

Overall, the quantitative analysis confirms that the classical optimization stage systematically reduces harmonic fragmentation, increases structural stability, and introduces a more balanced functional distribution, while maintaining acceptable melodic compatibility.

**Table 2.** Harmonic structure metrics.

| Variant | T | S | D | Other |
|---|---|---|---|---|
| Autumn Raw | 61,6% | 17,0% | 21,4% | 0,0% |
| Autumn Opt | 52,8% | 16,2% | 23,2% | 7,7% |
| Tipperary Raw | 73,5% | 11,8% | 14,7% | 0,0% |
| Tipperary Opt | 67,1% | 8,2% | 23,5% | 1,2% |

**Table 3.** Harmonic stability and bass movement.

| Song | Raw Avg Duration | Optimized Avg Duration | Raw vs Opt Bass Jump Δ |
|---|---|---|---|
| Autumn Leaves | 1,00 | 1,87 | 0,56 |
| It's a Long Way to Tipperary | 1,00 | 1,98 | 0,72 |

### 6.3 Interpretation and Expert Evaluation

The quantitative findings are partially supported by a qualitative evaluation involving two professional musicians and six non-expert listeners. While the sample size is limited, the evaluation provides initial insight into the perceived quality of the generated harmonizations.

For Autumn Leaves, the optimized version was consistently perceived as more coherent and structurally convincing than the raw harmonizer output. In particular, improvements were noted in cadential resolution and overall harmonic organization.

In contrast, for It's a Long Way to Tipperary, the expert evaluation revealed a different pattern. While the optimized version exhibited improved structural properties according to quantitative metrics, it was perceived as less natural in some respects, particularly due to increased harmonic complexity and less intuitive bass movement.

This divergence suggests that the effectiveness of the optimization stage depends on the stylistic characteristics of the input melody. For harmonically rich jazz standards such as Autumn Leaves, increased structural complexity may enhance musical plausibility. However, for simpler, more diatonic melodies such as Tipperary, excessive harmonic elaboration may reduce perceived naturalness.

These findings highlight an important aspect of harmonic decision-making: optimal solutions are context-dependent and cannot be fully captured by structural metrics alone. The interaction between harmonic complexity, stylistic expectations, and perceptual coherence plays a crucial role in evaluating generated harmonizations. One of the professional evaluators noted that at structurally important points of the melody, particularly at local peaks, harmonization should emphasize stronger functional roles. In this context, dominant chords were consistently perceived as providing stronger support than alternative harmonizations based on weaker functions (e.g., mediant), even when alternative solutions were theoretically valid.

## 7 Discussion

The proposed framework introduces a quantum-inspired perspective on harmonic generation, interpreting the process as decision-making within a structured combinatorial space. Rather than treating harmonization as purely statistical sequence prediction, the model explicitly incorporates constraints derived from tonal harmony and explores their interaction through an interference-based search mechanism.

The experimental results suggest that this hybrid approach captures several important aspects of human-like harmonic reasoning. In particular, the reduction of chord density and the increase in harmonic stability indicate that the classical optimization stage effectively refines the exploratory output of the harmonizer into more coherent structures. At the same time, the quantum-inspired stage enables the simultaneous consideration of multiple harmonic alternatives, which is difficult to achieve using purely deterministic or greedy methods.

From the perspective of cognitive modeling, the framework aligns with the view that human decision-making does not operate as a strictly sequential or deterministic process. Instead, multiple competing alternatives may coexist and be evaluated in parallel, with context-dependent factors influencing the final selection. The use of superposition-like representations and interference-inspired evaluation provides a computational analogy for such processes, without assuming any underlying physical quantum mechanisms. Such interpretations are consistent with broader efforts to apply quantum-theoretic formalisms to complex systems and cognition (Kitto, 2008; Khrennikov, 2013).

The observed differences between musical examples further highlight the context-dependent nature of harmonic decision-making. While the optimized harmonization improved perceived coherence in harmonically rich material (Autumn Leaves), it was less effective for simpler melodic structures (Tipperary). This suggests that the balance between harmonic complexity and perceptual clarity is not universal, but depends on stylistic and structural characteristics of the musical input. Such variability is consistent with findings in cognitive science, where decision strategies adapt to the structure of the problem space (Gigerenzer and Gaissmaier, 2011).

Importantly, the proposed model should be understood as a computational abstraction rather than a claim about biological implementation. The term “quantum-inspired” refers here to the use of mathematical principles—such as parallel representation of alternatives and interference-like evaluation—rather than to physical quantum processes in the brain. This distinction is crucial for maintaining conceptual clarity and scientific validity.

Beyond music generation, the framework may have broader implications for modeling decision-making in other structured creative domains. Problems involving multiple competing solutions under complex constraints—such as design, language generation, or planning—may benefit from similar hybrid approaches combining exploratory search with structured refinement.

At the same time, the present study has several limitations. The experimental evaluation is limited in scale and relies on a small number of musical examples and expert assessments. Future work should extend the evaluation to larger datasets, incorporate multiple expert evaluators, and explore more systematic comparisons with existing algorithmic approaches.

Despite these limitations, the results indicate that interference-based search mechanisms provide a promising direction for modeling structured decision processes in creative domains. The integration of such mechanisms with domain-specific knowledge may offer a useful framework for bridging computational modeling and cognitive theories of decision-making.

An important limitation of the present study concerns the use of well-known musical material. Both evaluated melodies (Autumn Leaves and It’s a Long Way to Tipperary) are widely recognized, which may influence subjective evaluation. Expert and non-expert listeners are likely to be biased by their prior knowledge of the original harmonic structure, which affects their perception of alternative harmonizations.

This effect was explicitly noted by one of the professional evaluators, who emphasized that even when attempting to ignore the original harmony, expectations remain influenced by learned musical patterns. In particular, it was observed that structurally strong harmonic functions (such as the dominant) are preferred at key melodic peaks, as they provide greater perceived support than weaker alternatives (e.g., mediant-based harmonizations).

To reduce such bias and improve objectivity, future work should include experiments with newly composed or previously unknown melodies. This would allow evaluation of the proposed model in a setting where listeners are not influenced by prior exposure, and would enable a clearer assessment of harmonic plausibility across both expert and non-expert groups.

Furthermore, the present evaluation involved a limited number of participants, including two expert musicians and a small group of non-expert listeners. While the results provide initial insights, larger-scale studies are needed to establish statistically robust conclusions.

# 8 Conclusion

This paper presented a quantum-inspired computational framework for harmonic decision-making in music. The proposed approach formulates harmonization as an optimization problem within a structured combinatorial space, where multiple candidate solutions are explored and evaluated under interacting constraints.

The framework combines two complementary components: an interference-based harmonization stage that generates candidate chord sequences, and a classical optimization stage that refines these candidates using established principles of tonal harmony. This hybrid architecture allows the model to balance exploratory search with structural coherence.

Experimental results demonstrate that the optimization stage systematically improves harmonic stability, reduces fragmentation, and produces more structured chord progressions. At the same time, the qualitative evaluation highlights the importance of stylistic context, showing that increased harmonic complexity is not universally beneficial and must be aligned with the character of the input melody.

From a broader perspective, the proposed model contributes to the understanding of creative decision-making as a process of navigating structured spaces of alternatives. The use of quantum-inspired concepts provides a flexible computational framework for representing and evaluating multiple competing solutions without relying on purely deterministic strategies.

Although preliminary, the results suggest that interference-based search mechanisms, when combined with domain-specific knowledge, may offer a promising direction for modeling complex decision processes in both music and other creative domains.

Future work will focus on extending the evaluation to larger datasets, incorporating multiple expert assessments, and exploring the applicability of the proposed framework to other forms of structured creative generation.

The implementation and evaluation data are publicly available to support reproducibility.

# Data availability statement

The data and code supporting the findings of this study are available in a public repository at: [https://github.com/JosefPavlicek/quantum-harmonic-decision-model.git].